%% file: iclr_arxiv.tex
\documentclass{article}
\usepackage{iclr2027_conference,times}

\input{math_commands.tex}

\usepackage{amsmath,amssymb,mathtools}
\usepackage{booktabs}
\usepackage{colortbl}
\usepackage{tabularx}
\usepackage{xcolor}
\usepackage{xspace}
\usepackage{microtype}
\usepackage{graphicx}
\usepackage{subcaption}
\usepackage{wrapfig}
\usepackage{float}
\usepackage[skins]{tcolorbox}
\usepackage{hyperref}
\definecolor{cyblue}{RGB}{0,173,238}
\hypersetup{colorlinks=true,linkcolor=black,citecolor=black,urlcolor=black}
\hypersetup{
	colorlinks=true,
	linkcolor=cyblue,
	citecolor=cyblue,
	urlcolor=cyblue
}
\usepackage{url}
\usepackage{enumitem}
\title{FACT: Feedback Attribution with Controlled Trajectories for Long-Horizon Artifact Agents}

\author{%
Lecheng Yan$^{1,2}$\quad
Jianze Lin$^{3}$\quad
Ben Pan$^{4}$\quad
Ruizhe Li$^{5}$\quad
Longyue Wang$^{6}$\\
\bfseries Wenxi Li$^{7}$\quad
Binwu Wang$^{1}$\quad
Chenyang Lyu$^{6}$\quad
Guanhua Chen$^{2}$\\[4pt]
\normalfont $^{1}$University of Science and Technology of China\quad
$^{2}$Southern University of Science and Technology\\
\normalfont $^{3}$Beijing Normal University\quad $^{4}$Tianjin University\quad
$^{5}$University of Birmingham\\
\normalfont $^{6}$Alibaba Group\quad $^{7}$East China Normal University
}

\definecolor{PromptBlue}{HTML}{4F6F8F}
\definecolor{PromptBlueBG}{HTML}{EEF3F7}
\definecolor{PromptSage}{HTML}{58766B}
\definecolor{PromptSageBG}{HTML}{EEF4F1}
\definecolor{PromptAmber}{HTML}{86693D}
\definecolor{PromptAmberBG}{HTML}{F7F2E8}
\newtcolorbox{promptbox}[2][]{
  enhanced,
  colback=white,
  colframe=PromptBlue!65!black,
  colbacktitle=PromptBlueBG,
  coltitle=PromptBlue!75!black,
  fonttitle=\bfseries,
  title={#2},
  boxrule=0.45pt,
  borderline west={1.5pt}{0pt}{PromptBlue},
  sharp corners,
  left=6pt,
  right=6pt,
  top=5pt,
  bottom=5pt,
  toptitle=3pt,
  bottomtitle=3pt,
  before skip=6pt,
  after skip=6pt,
  #1
}
\newtcbox{\promptaction}{on line,nobeforeafter,colback=PromptBlueBG,colframe=PromptBlue!35,boxrule=0.25pt,arc=1pt,boxsep=0pt,left=2pt,right=2pt,top=1pt,bottom=1pt,coltext=PromptBlue!80!black}
\newtcbox{\prompteval}{on line,nobeforeafter,colback=PromptSageBG,colframe=PromptSage!35,boxrule=0.25pt,arc=1pt,boxsep=0pt,left=2pt,right=2pt,top=1pt,bottom=1pt,coltext=PromptSage!80!black}
\newtcbox{\promptconstraint}{on line,nobeforeafter,colback=PromptAmberBG,colframe=PromptAmber!35,boxrule=0.25pt,arc=1pt,boxsep=0pt,left=2pt,right=2pt,top=1pt,bottom=1pt,coltext=PromptAmber!85!black}
\iclrfinalcopy
\begin{document}

\maketitle
\lhead{Preprint}

\begin{abstract}
Long-horizon artifact agents transform persistent environments through many interdependent tool actions before receiving feedback on the completed artifact. This delayed and often subjective feedback makes it difficult to determine which earlier decisions improved the outcome and to turn that evidence into reliable policy updates. 
PPO and group-relative policy optimization methods such as GRPO assign rollout-level or group-relative advantages to sampled sequences; delayed terminal feedback and stochastic tool execution leave action-level credit unresolved. 
We therefore introduce FACT, a critic-based PPO formulation that defines action credit by the expected effect of a committed action on terminal preference.
FACT executes alternative actions from the same restored workspace and continues each branch under a shared policy, yielding matched long-horizon comparisons. These comparisons supervise an action-conditioned transition critic for ordinary rollouts, while held-out interventions calibrate the resulting updates. We further introduce LongEditBench, a source-grounded benchmark for long-horizon video editing and contextual revision under follow-up user feedback. Experiments show that FACT improves completion-adjusted artifact quality and contextual revision, while controlled evaluations demonstrate more faithful action attribution. Consistent gains across transfer benchmarks further validate its effectiveness across persistent artifact environments.
\end{abstract}

\section{Introduction}
\label{sec:introduction}

Language-model agents increasingly produce persistent artifacts through tool-mediated workflows, including executable applications and long-form video editing \citep{trivedi2024appworld,yan2026crayottertraceable,cao2026agenticvbench}. 
We call these systems \textit{long-horizon artifact agents}: they use tools over multiple turns to create and revise persistent outputs.
Users inspect and revise these artifacts, so an output evolves through many state-changing actions. An early selection constrains later composition, a local revision can alter the coherence of the whole artifact, and user preference becomes apparent only after these choices have been assembled into an output \citep{yan2026crayotter,yan2026crayottertraceable}. 
A terminal judgment therefore leaves two linked uncertainties: which committed actions changed expected final quality, and whether a low outcome reflects the action or what happened during its execution.

Task-local preference comparisons provide a useful starting point for subjective artifact quality \citep{yan2026crayotter}. 
PPO uses a clipped actor update with a return or critic advantage, while GRPO and GSPO use relative group or sequence objectives; GRPB extends this family to long-horizon artifact workflows with rank-based credit \citep{shao2024deepseekmath,zheng2025gspo,yan2026crayotter}.
These signals remain attached to a rollout, group, or realized branch: PPO, GRPO, and GSPO leave action-level credit implicit, while GRPB can still assign a low score to useful actions when an export fails or an unexpected state appears. Existing policy signals thus conflate the quality of a committed action with execution variation.

\begin{wrapfigure}{r}{0.5\textwidth}
\vspace{-0.8\baselineskip}
\centering
\includegraphics[width=0.5\textwidth]{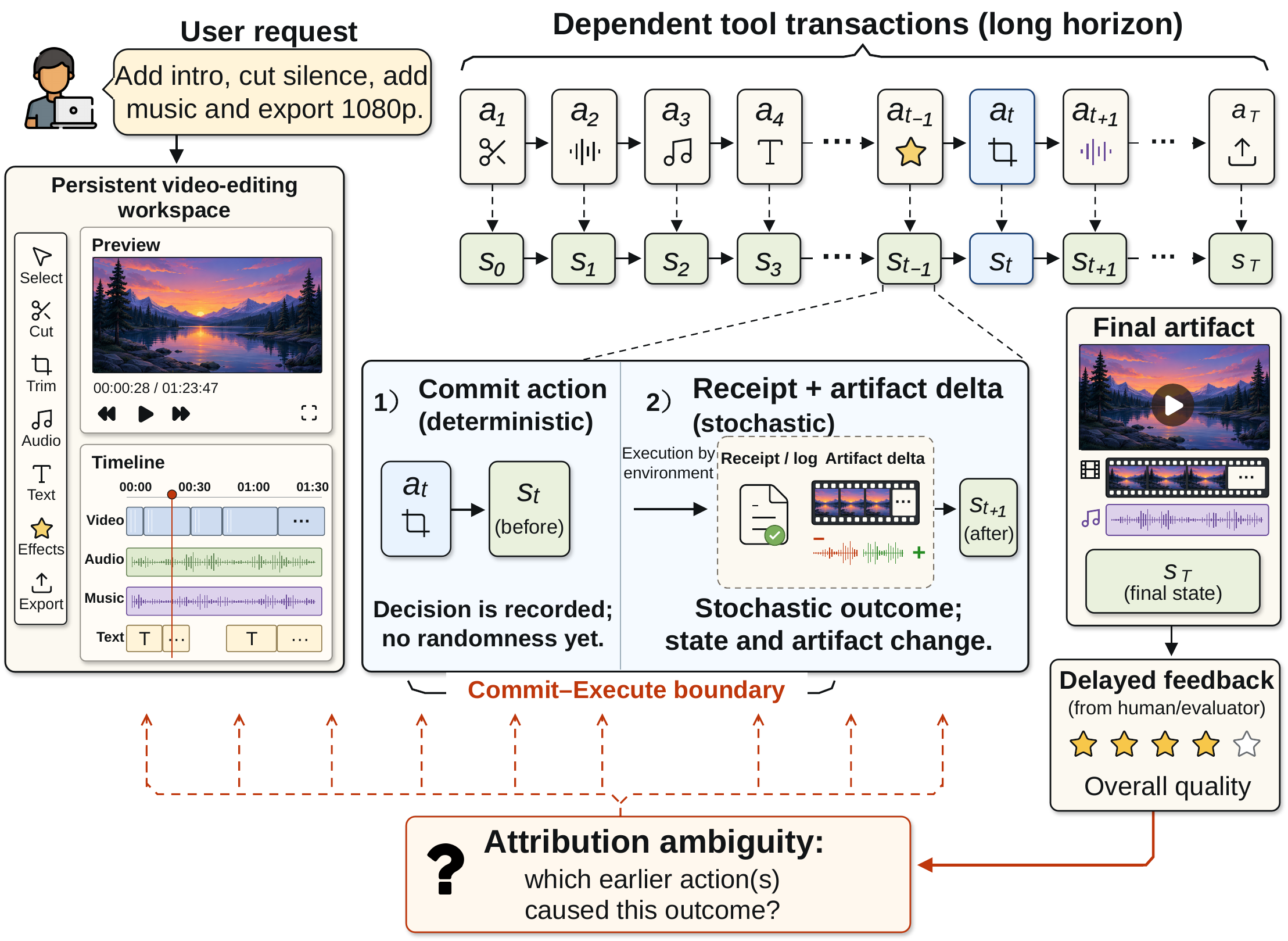}
\caption{Why rollout-level and group-relative advantages are insufficient for long-horizon artifact agents. A trajectory receives terminal feedback only after many state-changing actions; an action may be useful even when an unpredictable tool response or export failure lowers the realized outcome, motivating separate action and execution attribution.}
\label{fig:artifact_credit_problem}
\vspace{-0.6\baselineskip}
\end{wrapfigure}

These gaps motivate a local counterfactual question: \textit{how would expected final quality change if the agent committed to a different action from the same workspace, and how much of the observed outcome comes from the commitment versus its execution?} Figure~\ref{fig:artifact_credit_problem} illustrates the distinction. Counterfactual replay and intervention-based diagnosis use related comparisons \citep{shah2026car,zhang2026creditwithout}; fixing the preceding history and continuation policy makes the comparison local while preserving long-horizon consequences. Averaged continuations estimate the expected action effect, whereas the observed post-action outcome supplies an execution residual. This separation provides the two signals used by FACT for policy learning and diagnosis.

We develop this principle in FACT, \textit{Feedback Attribution with Controlled Trajectories}, as a critic-based PPO update. FACT keeps PPO's clipped actor and replaces its rollout advantage with calibrated action credit.
Controlled trajectories branch from a shared workspace and execute alternative actions under a fixed continuation policy; their expected terminal contrast is Agency credit. 
An action-conditioned transition critic transfers this branch supervision to ordinary rollouts, while the execution residual, Revelation, calibrates updates and supports diagnosis. Figure~\ref{fig:fact_overview} summarizes these components.

To evaluate action credit as artifacts and user requests evolve, we construct LongEditBench by manually adapting source-grounded interactions from LongShOTBench into executable editing briefs. The benchmark contains 337 requests over 21 videos, including 84 follow-ups \citep{kurpath2025longshot,yan2026crayottertraceable}. Its contextual follow-ups test incorporation of a new request while preserving earlier requirements. Transfer uses 100 AgenticVBench tasks across four families, 2,722 SpreadsheetBench Pass@1 tasks, and 61 held-out SpreadsheetBench 2 workflow tasks \citep{cao2026agenticvbench,ma2024spreadsheetbench,zhu2026spreadsheetbench2}. Matched-budget comparisons use a shared Qwen3.5 policy family and include several PPO variants, GRPO, GSPO, GRPB, and Direct-$Q$ for attribution \citep{shao2024deepseekmath,zheng2025gspo,yan2026crayotter}. Sibling interventions test attribution fidelity, while the transfer suites test policy utility across artifact domains. 

In summary, FACT defines action credit as the expected terminal effect of a committed action, separates that effect from execution residuals, and transfers controlled-branch supervision through an action-conditioned transition critic into a calibrated PPO update.
We introduce LongEditBench with contextual follow-ups and evaluate both attribution fidelity and policy utility across persistent artifact domains.
Results show gains in completion-adjusted artifact quality and multi-turn contextual revision, including follow-up instruction satisfaction and source grounding; attribution audits improve action-effect direction recovery, while cross-domain tests achieve the highest average artifact utility.


\section{Related Work}
\label{sec:related_work}

\paragraph{Temporal and counterfactual credit assignment}

Return redistribution and counterfactual advantage estimation address complementary aspects of delayed supervision. RUDDER learns contributions that redistribute trajectory return \citep{arjona2019rudder}, while Hindsight Credit Assignment and COCOA associate earlier actions with subsequent rewarding outcomes \citep{harutyunyan2019hca,meulemans2023cocoa}. Counterfactual Credit Assignment uses future-conditioned baselines to account for exogenous variation \citep{mesnard2021cca}; Direct Advantage Estimation and its off-policy extension estimate action-dependent contributions from return \citep{pan2022dae,pan2024skill}. GRPB studies long-horizon credit assignment in artifact workflows \citep{yan2026crayotter}, placing it alongside policy-optimization baselines in our comparisons. FACT makes this target operational for artifact agents through restored-workspace interventions and an 
action-conditioned transition critic that models the intervening tool response and artifact change.

\paragraph{Process supervision and credit for language agents}

Process supervision supplies intermediate correctness or progress signals for long reasoning trajectories \citep{uesato2022process,lightman2023verify,wang2023mathshepherd,cui2025prime}. For interactive agents, POAD connects inter-action credit to within-action token optimization \citep{wen2024poad}; Turn-PPO formalizes turn-level optimization \citep{li2026turnppo}; AgentPRM learns step-wise promise and progress \citep{xi2025agentprm}; GiGPO compares actions at repeated anchor states \citep{feng2025gigpo}; and hierarchical methods align supervision with subgoals and control structure \citep{he2026hgpo,peng2026hiper}. Progress-based estimators include $\Delta$Belief-RL, which measures changes in solution probability \citep{auzina2026deltabelief}, and TRACE, which derives telescoping turn rewards from frozen-reference answer likelihoods \citep{tao2026trace}. These approaches localize delayed supervision, whereas our setting requires estimating an action's expected effect on a persistent artifact by comparing commitments at a common workspace and modeling their subsequent consequences.

\paragraph{Executed attribution and artifact evaluation}

Trajectory diagnosis and intervention methods identify consequential steps, estimate effects through reruns, or test executable repairs \citep{zhang2025whowhen,zhang2025agentracer,shah2026car,lin2026reflect,bonagiri2026causalflow,zhang2026creditwithout}. FACT turns this evidence into amortized policy learning. Action-conditioned world models predict tool consequences, while ensembles and selective prediction estimate uncertainty \citep{yu2026rwml,gao2026websynthesis,lakshminarayanan2017ensembles,geifman2019selectivenet}. Persistent artifact benchmarks span restorable application state in AppWorld, traceable video editing in Crayotter, post-production tasks in AgenticVBench, long-video evidence seeking in LongShOTBench, and spreadsheet manipulation from test-case evaluation to end-to-end workflows in SpreadsheetBench and SpreadsheetBench 2 \citep{trivedi2024appworld,yan2026crayotter,yan2026crayottertraceable,cao2026agenticvbench,kurpath2025longshot,ma2024spreadsheetbench,zhu2026spreadsheetbench2}. LongEditBench adds source-grounded editing with contextual follow-up requests.

\section{Feedback Attribution with Controlled Trajectories}
\label{sec:method}

Long-horizon artifact agents must infer how tool actions transform persistent artifacts when post-action states and long-term consequences are revealed only after execution. FACT combines interaction history with visual and tool observations and controlled branches, then transfers branch supervision through an action-conditioned transition critic.
We formulate long-horizon artifact optimization as critic-based PPO. A state $s_t=(h_t,x_t)$ combines interaction history and persistent artifact state, while action $a_t$ commits a tool call. A complete trajectory is $\tau=(s_1,a_1,o_1,\ldots,s_T,a_T,o_T,x_{T+1})$, ending at artifact $x_{T+1}$ with terminal preference $R_c(\tau)$. FACT retains PPO's clipped actor update and replaces its rollout advantage with controlled action credit.

\subsection{Policy Update with Action Credit}
\label{sec:setup}

Video editing makes the long-horizon credit problem concrete. Early preparation constrains later composition, so the value of a local decision may emerge only after several dependent edits. 
We therefore start from PPO's clipped policy update and replace its advantage with action credit. PPO uses the ratio $r_t(\theta)=\pi_\theta(a_t\mid s_t)/\pi_{\mathrm{old}}(a_t\mid s_t)$ and estimated advantage $\widehat A_t$ in the clipped objective:

\begin{equation}
\mathcal L_{\mathrm{PPO}}(\theta)
=-\mathbb E_t\!\left[\min\!\left(r_t(\theta)\widehat A_t,
\operatorname{clip}\!\left(r_t(\theta),1-\epsilon,1+\epsilon\right)\widehat A_t\right)\right].
\label{eq:ppo_objective}
\end{equation}

FACT replaces PPO's rollout advantage with calibrated Agency credit estimated from controlled branches. For a fixed continuation policy $\pi$, $Q^\pi(s,a)=\mathbb E_\pi[R_c(\tau)\mid s,\operatorname{do}(a)]$ is the terminal preference after committing $a$, and $V^\pi(s)=\mathbb E_{a'\sim\pi(\cdot\mid s)}[Q^\pi(s,a')]$ is the pre-commitment baseline. The action credit is

\begin{equation}
A^\pi(s,a)=Q^\pi(s,a)-V^\pi(s).
\label{eq:action_contribution}
\end{equation}


The intervention $\operatorname{do}(a)$ commits $a$ from the current artifact state and lets $\pi$ continue to completion, so $Q^\pi$ retains the long-horizon consequences of the commitment. At iteration $m$, $\pi=\pi_{\mathrm{old}}^{(m)}$ is frozen during collection and scoring; the updated actor is denoted by $\pi_\theta$.
PPO derives action advantages from sampled rollouts, while group-relative optimizers such as GRPO and GSPO derive relative signals from sampled groups; FACT assigns the actor update to the committed tool-action span using counterfactual action credit.

\begin{figure*}[t]
\centering
\includegraphics[width=0.96\textwidth]{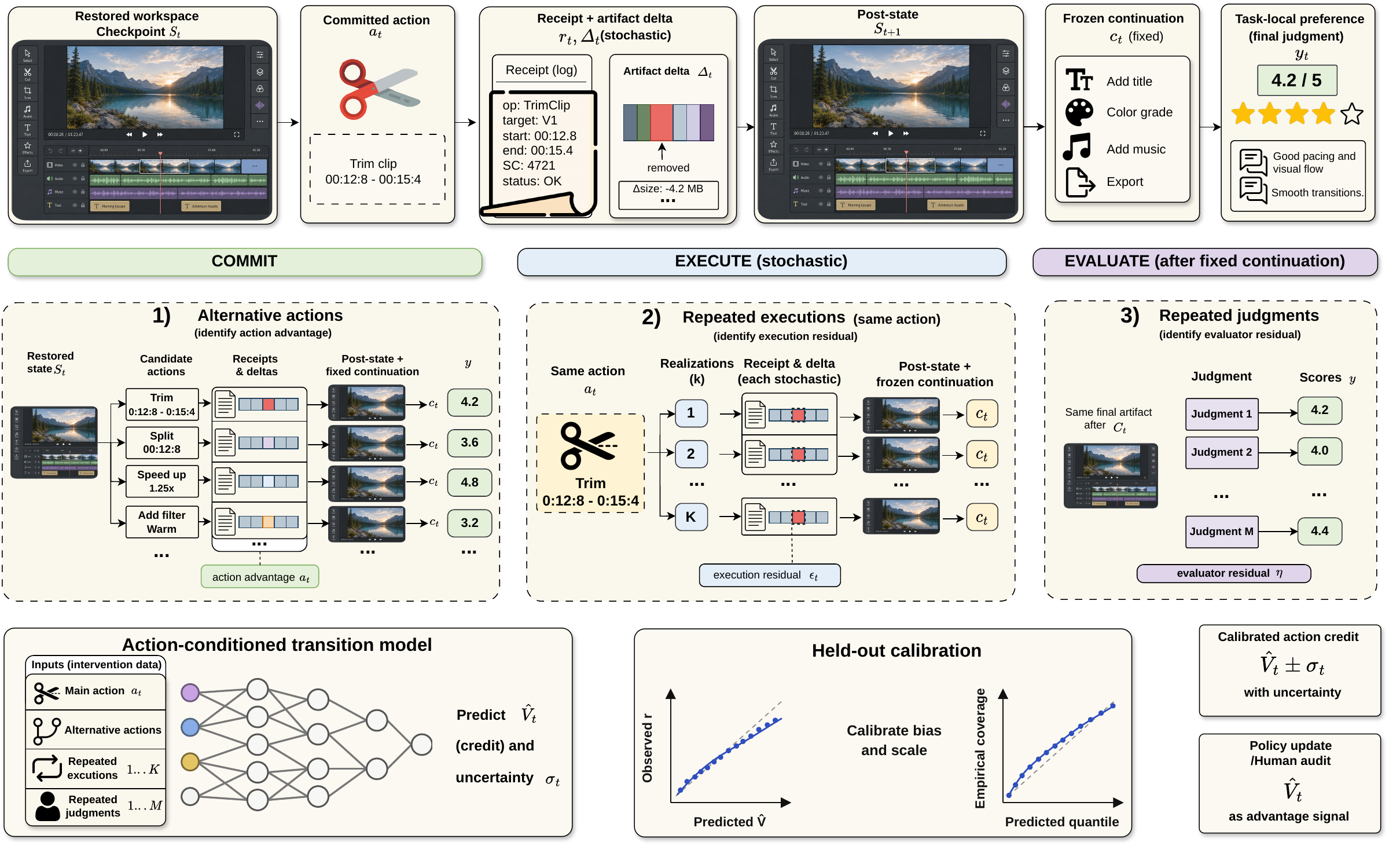}
\caption{FACT attributes final artifact feedback through controlled trajectories. Alternative actions executed from a shared artifact state supervise their relative contributions. An action-conditioned transition model predicts the response and artifact changes through which an action affects final quality. Held-out executed contrasts calibrate the credit used for policy updates, while recorded trajectories and branches support human inspection.}
\label{fig:fact_overview}
\end{figure*}

We score completed artifacts with task-local comparisons. For a frozen reference distribution $\mathcal D_c$ and signed preference score $\operatorname{pref}_c$, $R_c(\tau)=\mathbb E_{\tau'\sim\mathcal D_c}[\operatorname{pref}_c(x_{T+1},x'_{T+1})]\in[-1,1]$. The request-specific rubric places matched branches on a common outcome scale. FACT transfers their measured effects to ordinary trajectories and to the PPO objective in Section~\ref{sec:long_horizon_credit}. 
The bottom flow in Figure~\ref{fig:fact_overview}, from the branch observations through predicted credit and held-out calibration to the policy update, summarizes this transfer.

\subsection{Controlled branches and execution decomposition}
\label{sec:interventions}

In video editing, each stage produces a restorable project workspace that contains the timeline, source references, tool receipts, and current artifact. 
In the COMMIT panel of Figure~\ref{fig:fact_overview}, the restored state feeds candidate actions whose post-states are continued under the same frozen policy and scored at the common endpoint. This counterfactual replay follows prior intervention methods \citep{mesnard2021cca,meulemans2023cocoa,shah2026car}; fixing the workspace, tool schema, continuation, and endpoint isolates the committed action while preserving downstream effects.
Unlike group-relative ranking, the candidate contrast compares actions from the same workspace under the same continuation policy, isolating the expected effect of the commitment.

Repeated executions give $\bar R_i(a_{ik})=n_{ik}^{-1}\sum_{r=1}^{n_{ik}}R_{ikr}$ and the candidate-centered contrast

\begin{equation}
\widehat\Delta_i(a_{ik},a_{i\ell})=\bar R_i(a_{ik})-\bar R_i(a_{i\ell}).
\label{eq:branch_contrast}
\end{equation}


The contrast estimates $Q^\pi(s_i,a_{ik})-Q^\pi(s_i,a_{i\ell})$. The shared state-value baseline cancels, the common history fixes preceding decisions, and the continuation preserves downstream effects. 
Agency is the expected terminal effect of committing an action from a fixed workspace relative to the pre-commitment baseline. It is tied to an executable workspace intervention and can be tested on held-out sibling branches.


A committed action also encounters execution variability: the realized post-action state can differ across tool executions. Writing $A_t=A^\pi(s_t,a_t)$, the resulting value change decomposes into the expected contribution of the commitment and an execution residual:

\begin{equation}
\underbrace{V^\pi(s_{t+1})-V^\pi(s_t)}_{\text{realized value change}}=\underbrace{Q^\pi(s_t,a_t)-V^\pi(s_t)}_{A_t:\,\text{action contribution}}+\underbrace{V^\pi(s_{t+1})-Q^\pi(s_t,a_t)}_{\xi_t:\,\text{execution residual}}.
\label{eq:factorization}
\end{equation}

Under a fixed continuation policy, $\mathbb E[\xi_t\mid s_t,a_t]=0$. Revelation denotes this realized execution residual caused by the stochastic tool response and the resulting post-action state; repeated branches estimate it for calibration and inspection.




\subsection{Action-conditioned transition critic and calibrated PPO update}
\label{sec:estimator}
\label{sec:long_horizon_credit}

\paragraph{Action-conditioned transition critic.} 
In a long video edit, an early tool call creates a timeline and file state that later tools must read, while the post-action state is unavailable at decision time and can only be observed after the tool executes. Matched branches therefore supervise an action-conditioned transition critic that predicts tool consequences beyond selected workspaces, extending learned environment models \citep{yu2026rwml,gao2026websynthesis} with visual, tool, and interaction-history observations. The bottom of Figure~\ref{fig:fact_overview} show how these observations become predicted credit, uncertainty, and calibrated action credit before the \emph{Policy update / Human audit} block. We call the joint transition predictor and value head $(f_\phi,g_\omega)$ the action-conditioned transition critic.

We use $z_t=\psi(s_t)$ as the model representation of the history and persistent workspace, where $s_t=(h_t,x_t)$. The observation $o_t$ denotes the tool receipt and execution observation produced after committing $a_t$. The action representation $e(a_t)$ describes the tool, editing stage, and arguments. The action-conditioned transition critic predicts $(\widehat o_t,\widehat z_{t+1})=f_\phi(z_t,e(a_t))$, $\widehat Q_t=g_\omega(\widehat z_{t+1},\widehat o_t)$, $\widehat V_t^+=g_\omega(z_{t+1},o_t)$, $\widehat A_t=\widehat Q_t-V_\psi(z_t)$, and $\widehat\xi_t=\widehat V_t^+-\widehat Q_t$.

Each ensemble member predicts the transition and values both predicted and observed post-action states. The value function $V_\psi$ estimates value before commitment. Pre-execution $\widehat Q_t$ therefore identifies Agency, while post-execution $\widehat V_t^+$ identifies Revelation. The estimator jointly fits tool responses, workspace changes, terminal consequences, Agency contrasts, and Revelation residuals. Ensemble variation records uncertainty when credit is transferred to ordinary rollouts. 
Unlike Direct-$Q$, which predicts a scalar action value without modeling the intervening artifact transition, the transition critic predicts the post-action consequence and uses it to estimate action-conditioned credit.

The policy objective is $J(\theta)=\mathbb E_{\tau\sim\pi_\theta}[R_c(\tau)]$. FACT supplies the PPO advantage with critic-estimated action credit, so early choices retain their downstream contribution; the estimator is refreshed as the behavior policy changes.

\paragraph{Reliability calibration and PPO update.} 
Video stages differ in how visibly and reliably a tool changes the artifact, so advantage magnitudes are calibrated by stage.
 We form $u_t$ from ensemble disagreement and predicted value variance. Held-out restored-workspace contrasts fit its isotonic map to squared error within $g\in\{\text{selection},\text{construction},\text{finishing}\}$. Let $S_g$ be the weighted mean squared observed effect and $\operatorname{MSE}_g(u_t)$ the calibrated error. The resulting credit is
\begin{equation} \bar A_t=\frac{1}{B}\sum_{b=1}^{B}\widehat A_{t,b},\qquad w_t=\frac{S_g}{S_g+\operatorname{MSE}_g(u_t)},\qquad \widetilde A_t=w_t\bar A_t. \label{eq:calibrated_credit} \end{equation}
Larger held-out error produces a weaker update. Stage estimates are pooled toward the global curve according to calibration support.

Let $\mathcal I_t$ contain the tool-name and argument tokens that commit $a_t$. 
FACT applies PPO clipping to this span with calibrated Agency credit $\widetilde A_t$. Only calibrated Agency credit is supplied to the actor; Revelation is used for calibration and diagnosis.

Each iteration freezes the behavior policy, lagged estimator, and calibration map before computing credit. New trajectories and an independent intervention batch refresh them after the policy update. Tool responses remain context for later decisions and are masked from the loss.

For human inspection, training checkpoints retain complete trajectories, matched branches, attributed credit, and their artifact changes. Comparing these records over training shows which editing behaviors receive increasing reinforcement and whether they follow user intent. Full objective definitions and derivations are provided in Appendix~\ref{app:credit_details}.

\begin{figure}[t]
\centering
\includegraphics[width=0.98\textwidth]{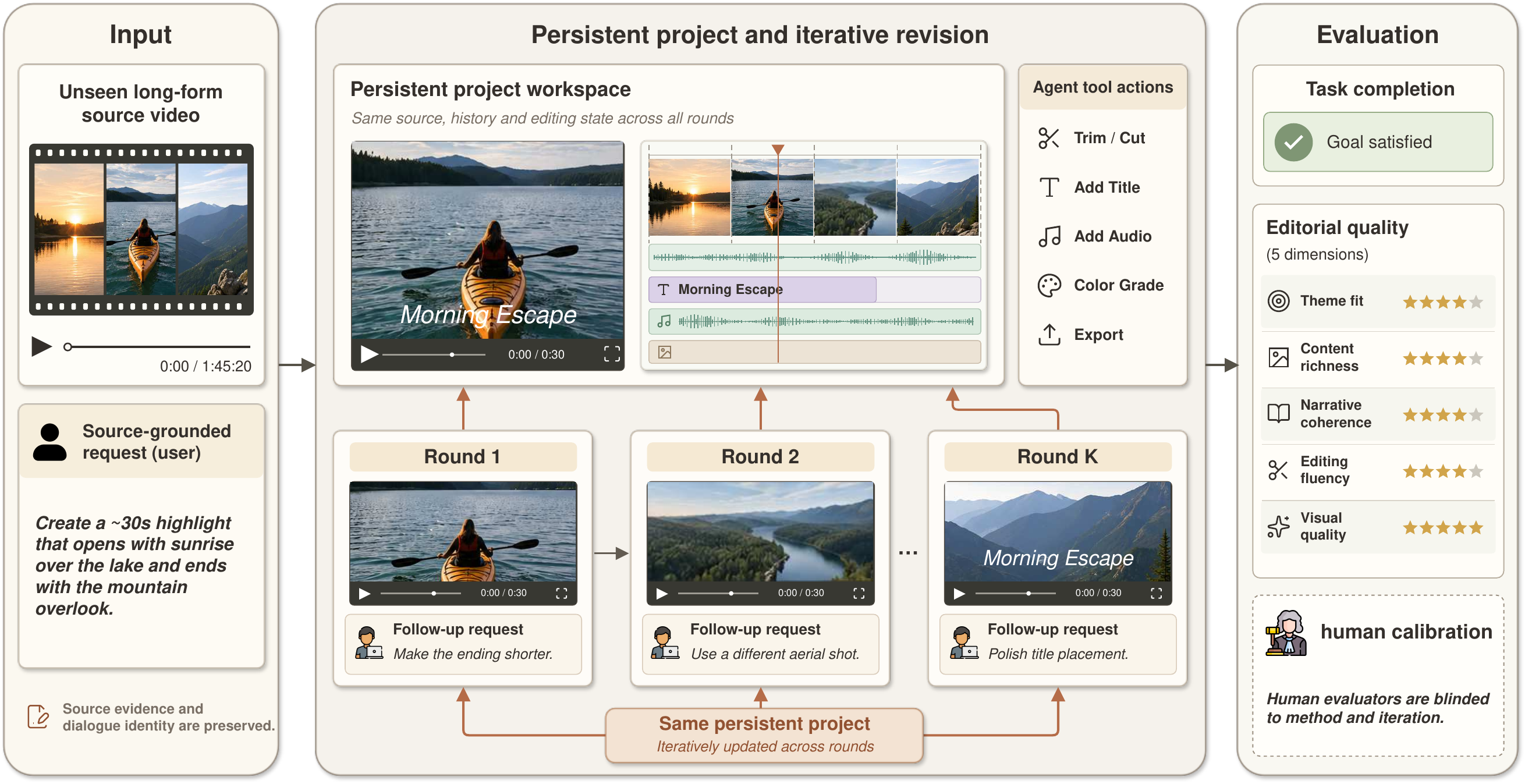}
\caption{LongEditBench protocol. A task couples evidence distributed across a long source with a persistent editing project; follow-up requests evaluate revision of the existing artifact under preserved source and task constraints.}
\label{fig:longedit_protocol}
\end{figure}

\section{Experiments}
\label{sec:evaluation}

We report the evaluation settings, LongEditBench protocol, policy-learning results, out-of-domain transfer, and interventional attribution.

\subsection{LongEditBench: long-horizon editing and contextual revision}
\label{sec:longedit_benchmark}

\paragraph{Capability and construction.} Existing long-video benchmarks evaluate source-grounded artifact production, while persistent artifact agents must also revise an existing project when a later request changes the target. LongEditBench defines this capability as contextual revision: after an initial edit, a follow-up request must be incorporated while preserving source evidence and applicable earlier constraints. We manually adapt source-grounded interactions from LongShOTBench into executable editorial briefs in Crayotter, yielding LongEditBench \citep{kurpath2025longshot,yan2026crayottertraceable}. It contains 337 requests over 21 videos, including 84 initial--follow-up pairs, and each request targets an approximately 30-second evidence-preserving edit.

\paragraph{Protocol and metrics.} For each pair, the follow-up receives the preceding interaction, rendered artifact, and retained project state, as shown in Figure~\ref{fig:longedit_protocol}; source access and applicable constraints persist, while reference answers and rubrics remain hidden. This tests evidence selection from a long source, persistent state maintenance, and revision under changed user intent. We report completion and five rubric dimensions: theme fit, content richness, narrative coherence, editing fluency, and visual quality. For valid-export indicator $c_i$ and five-point dimension scores $s_{id}$, the completion-adjusted score is
\begin{equation} S_{\mathrm{adj}}=\frac{1}{N}\sum_{i=1}^{N}c_i\left(\frac{1}{5}\sum_{d=1}^{5}s_{id}\right). \label{eq:completion_adjusted} \end{equation}
Revision additionally measures new-request satisfaction, preservation of applicable requirements, source grounding, and blinded preference; source-cluster bootstrap intervals account for shared videos.

\paragraph{Human audit.} The audit covers task validity, human reference quality, and the relation between rubric scores and human judgments. Two editors reviewed the converted cases with $27/42$ reviewer agreements (64.3\%). Three editors produced 27 human-reference videos with 100\% valid completion and mean $S_{\mathrm{adj}}=3.20/5$, indicating the difficulty of high-quality long-horizon editing. An independent three-grader panel scored 36 blinded outputs, obtaining human--evaluator Spearman $\rho=0.62$. Full task statistics, audit design, and disagreement analysis are provided in Appendices~\ref{app:longedit_stats}.

\begin{table*}[t]
\caption{LongEditBench evaluation. The upper panel reports completion and completion-adjusted five-point artifact quality over 337 requests (Equation~\ref{eq:completion_adjusted}); the lower panel reports contextual revision on 84 paired interactions. New req. scores the new follow-up requirement, Preserve scores applicable first-turn requirements that remain in force, Grounding scores source-evidence fidelity, Rev. avg. is the mean revision score, and Joint is the percentage of all pairs satisfying the new and preserved requirements together.}
\label{tab:longedit_results}
\centering
\small
\setlength{\tabcolsep}{2pt}
\renewcommand{\arraystretch}{1.08}
\begin{tabularx}{\textwidth}{@{}p{1.18in}*{6}{>{\centering\arraybackslash}X}@{}}
\toprule
& \multicolumn{2}{c}{2B} & \multicolumn{2}{c}{4B} & \multicolumn{2}{c}{9B} \\
\cmidrule(lr){2-3}\cmidrule(lr){4-5}\cmidrule(lr){6-7}
Method
& Comp. (\%) $\uparrow$ & $S_{\mathrm{adj}}\uparrow$
& Comp. (\%) $\uparrow$ & $S_{\mathrm{adj}}\uparrow$
& Comp. (\%) $\uparrow$ & $S_{\mathrm{adj}}\uparrow$ \\
\midrule
Base
& 91.1 & 1.845 & 95.8 & 1.939 & \textbf{98.5} & 2.091 \\
Terminal PPO
& 92.6 & 1.754 & 97.3 & 2.072 & 97.9 & 2.104 \\
Process PPO
& 92.9 & 1.739 & \textbf{98.8} & 2.096 & 97.9 & 2.082 \\
Terminal Rank PPO
& 92.6 & 1.725 & 97.6 & 2.045 & 97.3 & 2.058 \\
GRPO
& 92.6 & 1.748 & 97.6 & 2.072 & 97.3 & 2.114 \\
GSPO
& 94.1 & 1.776 & 98.2 & 2.114 & 97.6 & 2.129 \\
GRPB
& \textbf{95.0} & 1.926 & 96.7 & 1.992 & 97.6 & 2.099 \\
\rowcolor{PromptBlueBG}
FACT
& 93.8 & \textbf{1.975} & \textbf{98.8} & \textbf{2.119} & \textbf{98.5} & \textbf{2.210} \\
\bottomrule
\end{tabularx}
\vspace{3pt}
\begin{tabularx}{\textwidth}{@{}p{1.18in}*{5}{>{\centering\arraybackslash}X}@{}}
\multicolumn{6}{@{}l}{\textit{Contextual revision at 9B}} \\
\addlinespace[2pt]
\toprule
Method
& New req. $\uparrow$
& Preserve $\uparrow$
& Grounding $\uparrow$
& Joint (\%) $\uparrow$
& Rev. avg. $\uparrow$ \\
\midrule
Base
& 1.151 & 3.792 & 3.962 & 2.4 & 2.969 \\
Process PPO
& 1.148 & 3.754 & 4.131 & 2.4 & 3.011 \\
GSPO
& 1.214 & 3.786 & 4.161 & \textbf{4.8} & 3.054 \\
GRPB
& 1.203 & \textbf{3.847} & 3.932 & 3.6 & 2.994 \\
\rowcolor{PromptBlueBG}
FACT
& \textbf{1.283} & 3.800 & \textbf{4.183} & \textbf{4.8} & \textbf{3.089} \\
\bottomrule
\end{tabularx}
\end{table*}

\subsection{Experimental setup}

\paragraph{Tasks and metrics.} We separate artifact utility from attribution fidelity. On LongEditBench, completion is the fraction of valid checked exports, and $S_{\mathrm{adj}}$ is the completion-gated mean of five one-to-five rubric dimensions: theme fit, content richness, narrative coherence, editing fluency, and visual quality. For contextual revision, New req., Preserve, and Grounding score the new requirement, retained requirements, and source fidelity; Rev. avg. is their revision-quality summary, and Joint is the percentage of pairs satisfying the new and retained requirements together.

Attribution effects are centered within each executed sibling set. MSE and MAE measure squared and absolute centered-effect error, respectively; direction accuracy tests the sign of strong effects, rank correlation tests candidate ordering, and selection gain measures the utility of the highest-ranked action over uniform selection. SpreadsheetBench 2 additionally reports effect AURC, the risk--coverage area for uncertainty-selected predictions. Intervals are clustered by source or task family. Fixing the workspace, candidate set, continuation policy, and scorer makes the centered contrast isolate the effect of the committed action.

\paragraph{Comparisons.} We compare terminal, process, rank-based, group-relative, and segment-level PPO variants with FACT under matched training budgets. The comparison includes GRPO and GSPO as group-relative policy optimizers and GRPB as a long-horizon credit-assignment method for artifact workflows \citep{shao2024deepseekmath,zheng2025gspo,yan2026crayotter}. Direct-$Q$ replaces the action-conditioned transition critic with a scalar action-value predictor. Training, calibration, and test sources are disjoint; hyperparameters are listed in Appendix~\ref{sec:experimental_details}.

\paragraph{Budgets and uncertainty.} We report point estimates with source/task-cluster bootstrap intervals. Rollout and intervention budgets are matched within each comparison. Full rollout, intervention, calibration, and uncertainty settings are listed in Appendix~\ref{sec:experimental_details}.

\begin{table*}[t]
\caption{FACT ablations at 9B. Policy quality is measured on LongEditBench; attribution metrics use source-held-out executable sibling interventions. Effect MAE measures absolute centered-effect error, Dir. acc. measures sign agreement on strong effects, and Selection gain measures the executed utility of the selected sibling relative to uniform selection. Direct-$Q$ replaces the action-conditioned transition estimator with a scalar action-value predictor. ``W/o calibration'' removes continuous credit shrinkage while retaining the action-conditioned estimator.}
\label{tab:fact_ablation}
\centering
\small
\setlength{\tabcolsep}{2.5pt}
\renewcommand{\arraystretch}{1.16}
\resizebox{\textwidth}{!}{%
\begin{tabular}{@{}p{1.15in}
>{\centering\arraybackslash}p{0.65in}
>{\centering\arraybackslash}p{0.70in}
>{\centering\arraybackslash}p{0.74in}
>{\centering\arraybackslash}p{0.92in}
>{\centering\arraybackslash}p{0.72in}
>{\centering\arraybackslash}p{1.00in}@{}}
\toprule
Variant
& Estimator
& Calibration
& $S_{\mathrm{adj}}\uparrow$
& Effect MAE $\downarrow$
& Dir. acc. $\uparrow$
& Selection gain $\uparrow$ \\
\midrule
Direct-$Q$                & --         & \checkmark & 2.092    & 0.282    & 73.3\%   & \textbf{0.529} \\
FACT w/o calibration      & \checkmark & --         & 2.110    & \textbf{0.245}    & 50.0\%   & 0.086 \\
\rowcolor{PromptBlueBG}
FACT                      & \checkmark & \checkmark & \textbf{2.210}    & 0.258 & \textbf{92.3\%}   & \textbf{0.529} \\
\bottomrule
\end{tabular}%
}
\end{table*}

\subsection{Policy learning and core results}

\paragraph{LongEditBench results.}

Table~\ref{tab:longedit_results} reports frozen-policy quality over 337 requests and revision on 84 paired interactions. The upper panel measures valid artifact production; the lower panel tests new-request satisfaction, retention of earlier requirements, source grounding, and their joint success.
The complementary human audit reports expert task acceptance, a matched human reference, inter-rater reliability, and human--evaluator agreement in Table~\ref{tab:longedit_human_audit_results}. These measurements validate the benchmark and its scorer without turning human annotation into a second full evaluation matrix. Training dynamics are reported in Appendix~\ref{app:training_dynamics}.

\paragraph{Ablation.}

Table~\ref{tab:fact_ablation} isolates the action-conditioned estimator and credit calibration at 9B. $S_{\mathrm{adj}}$ measures artifact utility, Effect MAE and direction accuracy measure attribution fidelity, and Selection gain measures the utility of the selected action relative to uniform choice. All variants share initialization, intervention data, rollout budget, and optimizer.

\subsection{Out-of-domain generalization}
\label{sec:out_of_domain}

\paragraph{AgenticVBench}

Table~\ref{tab:agenticvbench_results} evaluates frozen 9B policies on AgenticVBench's repurposing, sequencing, repair, and assembly tasks \citep{cao2026agenticvbench}. Each score is the percentage of cases satisfying its post-production contract; Average is their arithmetic mean. The tasks cover complementary persistent-artifact operations under matched policies and checkpoints.

\begin{table*}[t]
\caption{AgenticVBench results (\%) for matched 9B policies. Average is the arithmetic mean of the four displayed task scores.}
\label{tab:agenticvbench_results}
\centering
\setlength{\tabcolsep}{7pt}
\begin{tabular}{lccccc}
\toprule
Method & Average $\uparrow$ & Repurpose $\uparrow$ & Sequencing $\uparrow$ & Repair $\uparrow$ & Assembly $\uparrow$ \\
\midrule
Base & 2.7 & 0.6 & 0.7 & 8.9 & 0.6 \\
Terminal PPO & 12.3 & 19.4 & 7.6 & 14.4 & 7.6 \\
Process PPO & 9.7 & 11.4 & 5.8 & 13.0 & 8.6 \\
Terminal Rank PPO & 9.8 & 12.5 & 6.7 & 13.4 & 6.4 \\
GRPO & 12.9 & 20.7 & 7.4 & 14.8 & 8.5 \\
GSPO & 12.4 & 22.3 & 8.0 & 14.2 & 5.2 \\
GRPB & 14.3 & 21.2 & 7.8 & \textbf{15.7} & 12.6 \\
\rowcolor{PromptBlueBG}
FACT & \textbf{15.2} & \textbf{22.4} & \textbf{8.3} & 15.3 & \textbf{14.7} \\
\bottomrule
\end{tabular}
\end{table*}

\paragraph{SpreadsheetBench}
\label{sec:spreadsheet_transfer}

\begin{table*}[t]
\caption{SpreadsheetBench 1 (left) and SpreadsheetBench 2 (right) policy results at 9B. SpreadsheetBench 1 reports official Pass@1 on 2,722 test tasks; SpreadsheetBench 2 reports percentages on 61 held-out tasks; $U_{\mathrm{sheet}}$ is the mean of Regression and Modification. Common SFT is the shared initialization; the SpreadsheetBench 2 RL rows use 100 updates. Submit denotes a submitted artifact, not task correctness.}
\label{tab:spreadsheet_policy}
\centering
\setlength{\tabcolsep}{4pt}
\resizebox{\linewidth}{!}{
\begin{tabular}{lcccccc}
\toprule
& \multicolumn{1}{c}{\textbf{SpreadsheetBench 1}}
& \multicolumn{5}{c}{\textbf{SpreadsheetBench 2}} \\
\cmidrule(lr){2-2}\cmidrule(lr){3-7}
Policy & Pass@1 $\uparrow$ & $U_{\mathrm{sheet}}\uparrow$ & Task acc. $\uparrow$ & Regression $\uparrow$ & Modification $\uparrow$ & Submit $\uparrow$ \\
\midrule
Common SFT & 7.94 & 22.58 & 1.64 & 34.96 & 10.20 & 36.07 \\
SFT + GRPO & 29.57 & 22.13 & 1.64 & 35.34 & 8.91 & 36.07 \\
SFT + GSPO & 30.27 & 23.53 & 1.64 & 36.89 & 10.16 & 37.70 \\
SFT + GRPB & 30.49 & 25.09 & 1.64 & 39.85 & 10.34 & \textbf{40.98} \\
\rowcolor{PromptBlueBG}
SFT + FACT & \textbf{32.33} & \textbf{26.13} & 1.64 & \textbf{39.88} & \textbf{12.37} & \textbf{40.98} \\
\bottomrule
\end{tabular}
}
\end{table*}

\begin{wraptable}{r}{0.43\textwidth}
\vspace{-0.7\baselineskip}
\caption{SpreadsheetBench 2 selective attribution. Errors are in $10^{-2}$ units.}
\label{tab:spreadsheet_attribution}
\centering
\footnotesize
\setlength{\tabcolsep}{2.6pt}
\renewcommand{\arraystretch}{1.10}
\resizebox{\linewidth}{!}{%
\begin{tabular}{@{}lcccc@{}}
\toprule
Estimator
& \shortstack{Effect\\AURC $\downarrow$}
& \shortstack{MAE\\@30\% $\downarrow$}
& \shortstack{MAE\\@50\% $\downarrow$}
& \shortstack{MAE\\@70\% $\downarrow$} \\
\midrule
Direct-$Q$ & 8.3 & 8.6 & 9.9 & 8.8 \\
\rowcolor{PromptBlueBG}
FACT & \textbf{8.1} & \textbf{7.0} & \textbf{8.9} & \textbf{8.6} \\
\bottomrule
\end{tabular}
}
\vspace{-0.6\baselineskip}
\end{wraptable}

We evaluate official SpreadsheetBench 1 Pass@1 on 2,722 test tasks and a 61-task workbook-family held-out fold of SpreadsheetBench 2 \citep{ma2024spreadsheetbench,zhu2026spreadsheetbench2}. For SpreadsheetBench 2, Task accuracy measures the requested operation, Regression preservation of existing behavior, Modification the requested change, and Submit valid artifact production; $U_{\mathrm{sheet}}$ averages Regression and Modification.

All SpreadsheetBench 2 conditions share a Qwen3.5-9B SFT checkpoint and 100 RL updates. Table~\ref{tab:spreadsheet_policy} reports final-workbook utility, and Table~\ref{tab:spreadsheet_attribution} reports selective attribution; training dynamics are provided in Appendix Figure~\ref{fig:spreadsheet_training_dynamics}. FACT leads in $U_{\mathrm{sheet}}$, Regression, and Modification, ties GRPB on Submit, and matches all RL policies on Task accuracy. The attribution estimator and calibration map are fixed before RL, with predictions and executed utilities centered within each restored-prefix candidate set.

\subsection{Interventional attribution}
\label{sec:agency_evaluation}

\begin{figure*}[t]
\centering
\begin{subfigure}[t]{0.235\textwidth}
\centering
\includegraphics[width=\linewidth]{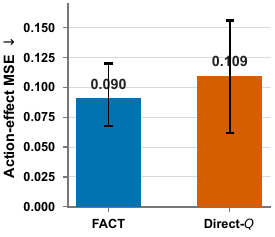}
\caption{Effect estimation.}
\label{fig:attribution_mse}
\end{subfigure}\hfill
\begin{subfigure}[t]{0.235\textwidth}
\centering
\includegraphics[width=\linewidth]{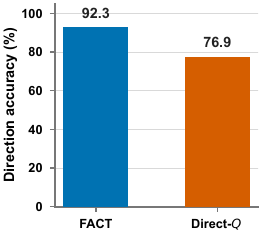}
\caption{Direction recovery.}
\label{fig:attribution_direction}
\end{subfigure}
\hfill
\begin{subfigure}[t]{0.235\textwidth}
\centering
\includegraphics[width=\linewidth]{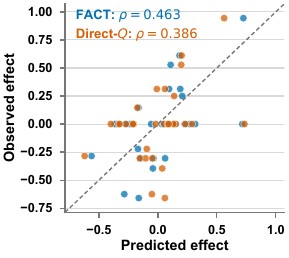}
\caption{Effect ranking fidelity.}
\label{fig:attribution_scatter}
\end{subfigure}\hfill
\begin{subfigure}[t]{0.235\textwidth}
\centering
\includegraphics[width=\linewidth]{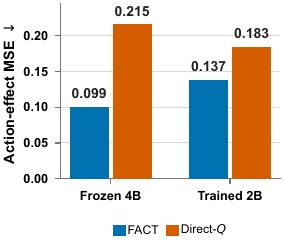}
\caption{Cross-scale transfer.}
\label{fig:attribution_cross_scale}
\end{subfigure}
\caption{Source-held-out action-effect attribution. Candidate actions are executed from matched workspaces and continued under a shared policy. The first three subfigures audit the frozen 9B policy; the last reports action-effect MSE for frozen 4B and trained 2B policies.}
\label{fig:attribution_results}
\end{figure*}

At each source-held-out prefix, three schema-valid candidate actions are executed from the same restored workspace and continued under the same frozen policy. Centering terminal utility within each candidate set makes MSE measure action-effect error and direction accuracy measure sign recovery for strong effects. The action-conditioned estimator reduces MSE relative to Direct-$Q$ from $0.215$ to $0.099$ in the frozen 4B audit and from $0.183$ to $0.137$ after 2B policy training, with direction-accuracy gains of $2.2$ and $9.7$ percentage points. Figure~\ref{fig:attribution_results} summarizes the attribution comparisons.

\section{Conclusion}

We presented FACT, a method for learning long-horizon artifact agents from delayed subjective feedback by attributing terminal preference to individual tool actions. FACT uses controlled trajectories from shared workspace states to measure the long-term effects of alternative actions, transfers this supervision through an 
action-conditioned transition critic
, and calibrates policy updates with held-out interventions. We also introduced LongEditBench to evaluate source-grounded artifact production and revision under follow-up user requests. Experiments across model scales, controlled attribution tests, and out-of-domain artifact environments show that FACT improves completed-artifact quality and contextual revision while producing action credits that better track executed effects. These results establish controlled trajectory attribution as a practical basis for optimizing and inspecting agents whose decisions accumulate into persistent artifacts.

\subsection*{AI use statement}

We used generative AI tools as writing and programming assistants. Their use included English-language polishing, revision of exposition, drafting and implementing experimental code, debugging, and formatting support. AI tools were not used to make scientific decisions, assign human-audit labels, or replace the reported evaluation procedures. The authors reviewed and verified all AI-assisted text, code, and figures and take responsibility for the final manuscript, experiments, and claims.

\subsection*{Ethics statement}

Human audits used independent raters who remained anonymous to one another. Raters did not receive one another's identities or judgments, and checkpoint identities, rewards, and execution traces were concealed. Only anonymized aggregate ratings are reported. Disagreement is retained as evidence of preference diversity rather than collapsed into a universal aesthetic target.

The training and evaluation records were assembled from author-curated task data, licensed or permissioned source resources, and benchmark-specific artifact records. Before the experiments, all source videos and related resources were manually checked for authorized use and known copyright conflicts. Resources with an identified ownership or licensing dispute were excluded, and no private personal information was intentionally collected for the benchmark or audit. Source materials were used only within the permitted research and evaluation scope; we do not knowingly redistribute material without the applicable authorization. The reported results contain anonymized measurements and aggregate artifact scores rather than identifiable rater information or restricted source files.

\subsection*{Reproducibility statement}

The LongEditBench dataset is available at \url{https://huggingface.co/datasets/witness0411/LongEditBench}.



\bibliography{iclr2027_conference}
\bibliographystyle{iclr2027_conference}

\appendix
\numberwithin{table}{section}
\numberwithin{figure}{section}

\section{Reproducibility Protocol}
\label{app:reproducibility}

Table~\ref{tab:fact_execution_protocol} fixes the order of data collection, estimation, calibration, and policy optimization. All quantities used by an actor update are computed before that update; newly acquired interventions refresh the estimator for the following iteration.

\begin{table}[htp]
\caption{FACT execution protocol. A source denotes a video or persistent project and remains within one split. The intervention-training, Agency-calibration, and realization-calibration sources are mutually disjoint.}
\label{tab:fact_execution_protocol}
\centering
\setlength{\tabcolsep}{4pt}
\renewcommand{\arraystretch}{1.12}
\begin{tabularx}{\textwidth}{@{}p{0.12\textwidth}p{0.20\textwidth}p{0.36\textwidth}X@{}}
\toprule
\textbf{Stage} & \textbf{Held fixed} & \textbf{Operation} & \textbf{Recorded output} \\
\midrule
Policy rollout & $\pi_{\mathrm{old}}$, lagged estimator, calibration map, evaluator & Execute the current policy to termination in an isolated workspace; score the exported artifact. & Complete trajectory, action-token spans, workspace snapshots, terminal utility \\
Candidate proposal & Prefix $s_i$, $\pi_{\mathrm{old}}$, tool schemas & Retain the committed action and sample schema-valid alternatives without replacement before observing branch outcomes. & Fixed candidate set $\mathcal A_i$ and proposal probabilities \\
Agency branches & $s_i$, $\mathcal A_i$, continuation policy, evaluator & Restore the same pre-action workspace, execute each candidate independently, and continue to the common endpoint. & Candidate artifacts, tool receipts, returns $R_{ikr}$, action contrasts $\widehat\Delta_i$ \\
Realization branches & $s_i$, action $a_{ik}$, continuation policy, evaluator & Repeat the same action, preserve each realized post-state, and run matched continuations from that post-state. & Execution-conditioned returns and residual target $\xi^\star_{ike}$ \\
Calibration & Frozen estimator; held-out sources & Compare predicted and executed candidate-centered effects; fit stage-pooled uncertainty--risk maps. & Calibrated error $\operatorname{MSE}_g(u)$ and continuous weight $w_t$ \\
Actor update & $\pi_{\mathrm{old}}$, estimator, calibration map & Apply the clipped objective only to the tokens that jointly encode the committed tool action. & Updated policy $\pi_\theta$ \\
Refresh & Updated policy; split assignment & Acquire an independent intervention batch and update the estimator and calibration state. & Frozen lagged state for the next actor update \\
\bottomrule
\end{tabularx}
\end{table}

\subsection{Prompt and Output Contracts}
\label{app:prompts}

Braced fields are populated at runtime, and tool definitions are supplied through the model's native function schema. Figure~\ref{fig:rollout_prompt} gives the policy prompt; Figure~\ref{fig:fact_auxiliary_prompts} gives the two constrained prompts used for alternative actions and terminal evaluation.

\begin{figure}[H]
\centering
\begin{promptbox}{Policy-rollout prompt template}
\raggedright
\textbf{System.} You are the Phase 3 execution agent for Crayotter. Convert the user's editing request and existing artifacts into a new, verified final video by calling the provided tools. Work as a persistent editing process: diagnose the prior version, preserve requested content, choose reusable material, build or revise the timeline, export, inspect, and repair when needed. \promptaction{Tool schemas are authoritative.} Use real paths returned by tools and never invent or rename files. Run up to two independent inspections or cuts together; otherwise read each observation before the next dependent action. Do not merely describe edits. A successful task ends only after a newly produced final artifact is \promptconstraint{exported and checked}. Technical failure should be repaired locally instead of restarting valid work.\par\smallskip
\texttt{Workspace: \{workspace\}}\quad \texttt{User materials: \{user\_workspace\}}\quad \texttt{Reference memory: \{memory\}}\par\smallskip
\textbf{User.} \texttt{\{user request\}} $\rightarrow$ \texttt{\{target duration\}} $\rightarrow$ \texttt{\{optional editing blueprint\}} $\rightarrow$ \texttt{\{tool-call protocol and schemas\}} $\rightarrow$ \texttt{\{initial-edit or revision contract\}} $\rightarrow$ \texttt{\{case metadata\}} $\rightarrow$ \texttt{\{workspace files and available source analysis\}}.\par\smallskip
\textbf{Required completion.} \promptaction{Produce a new final artifact}, export it, inspect the exported file, and repair a failed dependent operation from its returned observation.
\end{promptbox}
\caption{Prompt template for ordinary rollouts and controlled continuations. The initial-edit contract binds the source material; the revision contract additionally supplies the previous artifact, retained project state, and follow-up request. Instance data replace the braced fields without method-specific prompt rewriting.}
\label{fig:rollout_prompt}
\end{figure}

\begin{figure}[htp]
\centering
\begin{promptbox}{Alternative-action proposal}
\raggedright
\textbf{Input.} The complete tokenized trajectory prefix ending immediately before the committed tool action.\par\smallskip
\textbf{Tool selection suffix.}\par
\texttt{Choose the most plausible alternative next tool:}\par
Output is constrained to an unselected member of the active tool set.\par\smallskip
\textbf{Argument suffix.}\par
\texttt{Return only the grounded JSON arguments for counterfactual tool \{tool name\}:}\par
Output is constrained by that tool's JSON schema; additional properties are disabled. Tool choice and arguments are sampled from the frozen behavior policy. Invalid JSON is retried at most three times.
\end{promptbox}
\begin{promptbox}[colframe=PromptSage!65!black,colbacktitle=PromptSageBG,coltitle=PromptSage!80!black,borderline west={1.5pt}{0pt}{PromptSage}]{Frozen terminal evaluator}
\raggedright
You are the frozen product evaluator for a video-editing agent. \prompteval{Evaluate only the final video} against the user request; \promptconstraint{ignore tool calls and hidden reasoning}. Return \promptaction{exactly one JSON object} with keys \texttt{score}, \texttt{dimensions}, \texttt{hard\_failures}, and \texttt{critique}. The score is in $[0,100]$. Dimensions are \texttt{instruction\_following}, \texttt{narrative\_coherence}, \texttt{visual\_continuity}, \texttt{temporal\_structure}, \texttt{visual\_quality}, and \texttt{revision\_fidelity}, each in $[0,10]$. List at most four short hard failures and limit the critique to two short sentences. Return no Markdown.\par\smallskip
\texttt{User request: \{request\}}\par
\texttt{Target duration: \{seconds\}}
\end{promptbox}
\caption{Auxiliary prompt contracts. Alternative proposals use schema-constrained decoding from the same frozen policy; terminal evaluation receives the completed video and fixed task fields.}
\label{fig:fact_auxiliary_prompts}
\end{figure}

\begin{table}[htp]
\caption{Model-visible fields and their roles in the rollout prompt. Scorer references, held-out action effects, evaluator outputs, and calibration errors are excluded.}
\label{tab:prompt_fields}
\centering
\begin{tabularx}{\textwidth}{@{}p{0.20\textwidth}p{0.30\textwidth}X@{}}
\toprule
\textbf{Prompt block} & \textbf{Contents} & \textbf{Function} \\
\midrule
Task brief & Current user request, target duration, optional editing blueprint & Defines the requested artifact and constraints \\
Persistent context & Previous artifact and retained workspace for a follow-up turn & Supports diagnosis, preservation, and targeted revision \\
Environment state & Available files, paths, source analyses, and current artifact measurements & Grounds actions in the actual workspace \\
Tool interface & Active tool names, argument schemas, and call syntax & Constrains executable actions \\
Execution contract & Initial-edit or contextual-revision stage requirements & Specifies the completion sequence and final-artifact check \\
Reference memory & Training-fold workflow memory when available & Supplies reusable editing experience without benchmark answers \\
\bottomrule
\end{tabularx}
\end{table}

\section{Attribution and Calibration Details}
\label{app:credit_details}

For a fixed request and continuation policy, the tower property gives $\mathbb E[V^\pi(s_{t+1})\mid s_t,a_t]=Q^\pi(s_t,a_t)$. Here $A_t$ abbreviates $A^\pi(s_t,a_t)$ and $\xi_t$ denotes its corresponding execution residual; the terminal convention is $V^\pi(s_{T+1})=R_c(\tau)$. The population value $V^\pi$ is distinct from the learned pre-commitment estimator $V_\psi(z_t)$ used by the transition critic. Thus $\mathbb E[\xi_t\mid s_t,a_t]=0$, $\mathbb E_{a_t\sim\pi}[A_t\mid s_t]=0$, and Equation~\ref{eq:factorization} telescopes to $R_c-V^\pi(s_1)=\sum_t(A_t+\xi_t)$. A noisy terminal judgment $\widetilde R_c=R_c+\eta$ adds evaluator residual $\eta$. Conditioning the likelihood-ratio gradient on $(s_t,a_t)$ and subtracting the zero-mean baseline $V^\pi(s_t)$ yields Equation~\ref{eq:long_horizon_gradient}. These identities assume fixed request and environment distributions, a fixed evaluator, finite expected horizon, and differentiable policy probabilities.

The intervention horizon determines the scope of the resulting credit. Branches continued to task termination estimate the terminal action contrasts in Equation~\ref{eq:branch_target}. A branch evaluated at an earlier cutoff measures the effect under that cutoff and scoring rule. A truncated continuation can target terminal value through an explicit continuation-value estimate, whose error becomes part of attribution error. Each audit therefore specifies its continuation horizon and endpoint; held-out long-horizon validation uses the corresponding terminal target.

\begin{equation}
  \bar R_i(a_{ik})=\frac{1}{n_{ik}}\sum_{r=1}^{n_{ik}}R_{ikr},\qquad \widehat\Delta_i(a_{ik},a_{i\ell})=\bar R_i(a_{ik})-\bar R_i(a_{i\ell}).
  \label{eq:branch_target}
\end{equation}

\begin{equation}
  \begin{aligned}
  (\widehat o_t,\widehat z_{t+1})&=f_\phi(z_t,e(a_t)),&\widehat Q_t&=g_\omega(\widehat z_{t+1},\widehat o_t),\\
  \widehat V_t^+&=g_\omega(z_{t+1},o_t),&\widehat A_t&=\widehat Q_t-V_\psi(z_t),\qquad \widehat\xi_t=\widehat V_t^+-\widehat Q_t.
  \end{aligned}
  \label{eq:transition_critic}
\end{equation}

For checkpoint $i$ and action $k$, execution $e$ produces post-state $s'_{ike}$. This branch-specific post-action state is the realized counterpart of $s_{t+1}$ in the main text, with the execution observation included in the updated history. Averaging $J$ continuations from that state and centering over the execution mixture yields
\begin{equation} \bar R_{ike}=\frac{1}{J}\sum_{j=1}^{J}R_{ikej},\qquad \xi^\star_{ike}=\bar R_{ike}-\sum_{e'=1}^{E}w_{e'}\bar R_{ike'},\qquad \sum_{e'=1}^{E}w_{e'}=1. \label{eq:execution_target} \end{equation}
The centered residual target $\xi^\star_{ike}$ is the empirical Revelation target in the main text. The weights reflect execution frequencies, and matched continuations couple controllable randomness across post-states. Table~\ref{tab:variation_controls} separates the empirical controls.

In the action-contrast loss below, $\widehat Q_i(a_{ik})$ abbreviates the transition-critic prediction obtained from checkpoint $s_i$ and candidate action $a_{ik}$.

\begin{equation}
  \begin{aligned}
  \mathcal L_{\mathrm{sup}}&=\lambda_z\mathcal L_{\mathrm{next}}+\lambda_o\mathcal L_{\mathrm{resp}}+\lambda_R\mathcal L_{\mathrm{terminal}}+\lambda_V\mathcal L_{\mathrm{baseline}},\\
  \mathcal L_{\mathrm{act}}&=\mathbb E_{i,k,\ell}\!\left[\left(\widehat Q_i(a_{ik})-\widehat Q_i(a_{i\ell})-\widehat\Delta_i(a_{ik},a_{i\ell})\right)^2\right],\\
  \mathcal L_{\mathrm{FACT}}&=\mathcal L_{\mathrm{sup}}+\lambda_{\mathrm{act}}\mathcal L_{\mathrm{act}}+\lambda_{\mathrm{exec}}\mathbb E\!\left[\operatorname{Huber}(\widehat\xi-\xi^\star)\right].
  \end{aligned}
  \label{eq:estimator_loss}
\end{equation}

\begin{table}[H]
\caption{Intervention controls for action contribution, execution residual, and evaluator variation.}
\label{tab:variation_controls}
\centering
\begin{tabularx}{\textwidth}{@{}p{0.19\textwidth}p{0.27\textwidth}p{0.25\textwidth}X@{}}
\toprule
\textbf{Quantity} & \textbf{Held fixed} & \textbf{Varied} & \textbf{Estimate} \\
\midrule
Action contribution & Pre-action workspace, candidate set, continuation policy, endpoint, evaluator & Committed candidate action & Candidate-centered terminal return or pairwise action contrast \\
Execution residual & Pre-action workspace, fixed action, continuation policy, endpoint, evaluator & Tool execution outcome and resulting post-state & Return from one post-state relative to that action's execution-averaged return \\
Evaluator variation & Completed artifact and task brief & Independent evaluator draw & Difference between repeated judgments of the same artifact \\
Nominal replay floor & Workspace, action, execution seed, continuation seed, endpoint, evaluator seed & Independent replay of the identical specification & Residual score difference under nominally identical execution \\
\bottomrule
\end{tabularx}
\end{table}

For ensemble member $b$, let $\widehat\sigma^2_{Q,t,b}$ and $\widehat\sigma^2_{V,t,b}$ be predicted value variances. The calibration input is
\begin{equation} u_t^2=\operatorname{Var}_b(\widehat A_{t,b})+\frac{1}{B}\sum_{b=1}^{B}\left(\widehat\sigma^2_{Q,t,b}+\widehat\sigma^2_{V,t,b}\right). \label{eq:credit_uncertainty} \end{equation}
Held-out effects and errors are centered over each candidate set. Calibration uses inverse source--stage cell counts as weights and isotonic regression to map uncertainty to squared contrast error. Signal power is the weighted mean squared observed effect. Stage-specific risk and signal estimates are pooled toward their global counterparts according to sample count. Sparse calibration evidence blends risk toward a preset error level; uncertainty beyond calibrated support receives at least that risk. Equation~\ref{eq:calibrated_credit} uses these pooled quantities. Candidate-centered contrasts evaluate relative action effects, while the actor's state-value baseline is learned from behavior-policy returns.

\begin{equation}
  \nabla_\theta J(\theta)=\mathbb E_{\tau\sim\pi_\theta}\!\left[\sum_{t=1}^{T}\nabla_\theta\log\pi_\theta(a_t\mid s_t)\,A^{\pi_\theta}(s_t,a_t)\right].
  \label{eq:long_horizon_gradient}
\end{equation}

In FACT, the theoretical advantage $A^{\pi_\theta}$ in this identity is replaced in the clipped actor loss by the calibrated Agency estimate $\widetilde A_t$.

\begin{equation}
  \begin{aligned}
  \log\pi_\theta(a_t\mid s_t)&=\sum_{j\in\mathcal I_t}\log\pi_\theta(y_j\mid y_{<j},s_t),\\
  r_t(\theta)&=\frac{\pi_\theta(a_t\mid s_t)}{\pi_{\mathrm{old}}(a_t\mid s_t)},\\
  \mathcal L_{\mathrm{actor}}(\theta)&=-\mathbb E_t\!\left[\min\!\left(r_t(\theta)\widetilde A_t,\operatorname{clip}(r_t(\theta),1-\epsilon,1+\epsilon)\widetilde A_t\right)\right].
  \end{aligned}
  \label{eq:actor_loss}
\end{equation}

\section{Experimental Details}
\label{sec:experimental_details}

\subsection{Training Hyperparameters}

Tables~\ref{tab:actor_hyperparameters} and \ref{tab:fact_hyperparameters} separate shared policy optimization from FACT-specific acquisition, estimation, and calibration.

\begin{table}[H]
\caption{Shared policy-training and rollout hyperparameters. Experiments use Qwen3.5 at 2B, 4B, and 9B; the primary attribution audit and mechanism plots use 9B. Result-specific checkpoint deviations are stated with the corresponding table.}
\label{tab:actor_hyperparameters}
\centering
\begin{tabularx}{\textwidth}{@{}p{0.18\textwidth}p{0.35\textwidth}X@{}}
\toprule
\textbf{Category} & \textbf{Hyperparameter} & \textbf{Value} \\
\midrule
Model & Base policies & Qwen3.5-2B / 4B / 9B \\
& Parameter update & Full-parameter \\
& Training / rollout precision & bfloat16 / bfloat16 \\
& Maximum context length & 32,768 tokens \\
& Maximum prompt / response length & 16,384 / 16,384 tokens \\
\midrule
Rollout & Engine and mode & vLLM, asynchronous \\
& Prompt batch size & 4 \\
& Samples per prompt & 4 \\
& Maximum assistant turns & 50 \\
& Maximum tool-response length & 512 tokens \\
& Sampling temperature / top-$p$ & 0.8 / 0.95 \\
\midrule
Optimization & Training updates & 100 unless stated otherwise \\
& Actor learning rate & $1\times10^{-6}$ \\
& PPO minibatch / microbatch per GPU & 4 / 1 \\
& Policy ratio clip & 0.2 \\
& KL loss type / coefficient & low-variance KL / 0.001 \\
& Entropy coefficient & 0 \\
& Policy-loss unit & Joint tool-action token span \\
& Random seed & 2027 \\
\bottomrule
\end{tabularx}
\end{table}

\begin{figure}[H]
\centering
\includegraphics[width=\textwidth]{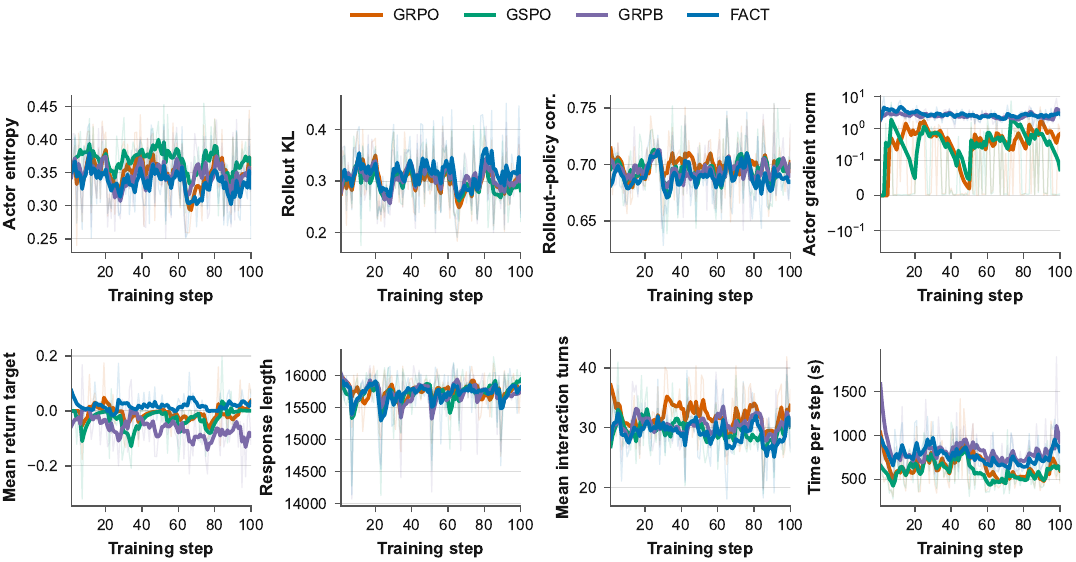}
\caption{SpreadsheetBench 2 training dynamics for the four 9B policy conditions in Table~\ref{tab:spreadsheet_policy}. The first row reports optimization diagnostics (actor entropy, rollout KL, rollout--policy correlation, and actor gradient norm); the second row reports shared training-side performance signals (mean return target, non-aborted response length, interaction turns, and time per update). Thin curves show per-update measurements and thick curves show seven-update EWMAs. All runs share the same SFT initialization, seed, sampling budget, and 100-update schedule.}
\label{fig:spreadsheet_training_dynamics}
\end{figure}

\begin{table}[H]
\caption{FACT-specific intervention acquisition, action-conditioned estimator, and reliability-calibration hyperparameters for full runs.}
\label{tab:fact_hyperparameters}
\centering
\begin{tabularx}{\textwidth}{@{}p{0.18\textwidth}p{0.42\textwidth}X@{}}
\toprule
\textbf{Category} & \textbf{Hyperparameter} & \textbf{Value} \\
\midrule
Interventions & Alternative actions per restored prefix & 3 \\
& Candidate-action temperature / top-$p$ & 0.8 / 0.95 \\
& Repeats per training / calibration action & 2 / 4 \\
& Executions / nested continuations per execution & 4 / 3 \\
& Anchor repeats for realization targets & 2 \\
& Branch prefixes per estimator update & 16 \\
& Online bootstrap rounds & 12 \\
& Training / Agency-calibration / realization windows & 64 / 128 / 8 prefixes \\
\midrule
Estimator & Ensemble members & 3 \\
& Hidden width & 128 \\
& Optimizer & AdamW \\
& Learning rate / weight decay & $10^{-3}$ / $10^{-3}$ \\
& Initial fit / online-update epochs & 60 / 4 \\
& Replay capacity / update batch size & 4096 / 1024 transactions \\
& Action-contrast loss weight & 4.0 \\
& Gradient-norm clipping & 5.0 \\
\midrule
Reliability & Minimum Agency / realization calibration points & 36 / 12 \\
& Minimum calibration sources / Agency strata & 5 / 3 \\
& Minimum selective-attribution coverage & 0.20 \\
& Calibration risk quantile & 0.80 \\
& Maximum accepted contrast error & 0.25 \\
& Minimum uncertainty--error Spearman correlation & 0.10 \\
\midrule
Evaluator & Decoding temperature / top-$p$ & 0.0 / 1.0 \\
& Maximum output length & 2,048 tokens \\
& Output constraint & Fixed JSON schema \\
& Score / dimension ranges & $[0,100]$ / $[0,10]$ \\
\bottomrule
\end{tabularx}
\end{table}

\section{Additional Training Dynamics}
\label{app:training_dynamics}

\begin{figure*}[t]
\centering
\captionsetup[subfigure]{font={bf,footnotesize}}
\begin{subfigure}[t]{0.24\textwidth}\centering\includegraphics[width=\linewidth]{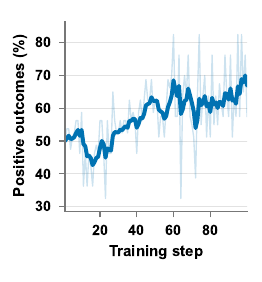}\caption{Positive outcome.}\end{subfigure}\hfill
\begin{subfigure}[t]{0.24\textwidth}\centering\includegraphics[width=\linewidth]{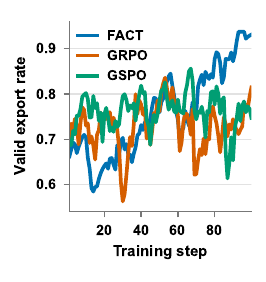}\caption{Executable outcome.}\end{subfigure}\hfill
\begin{subfigure}[t]{0.24\textwidth}\centering\includegraphics[width=\linewidth]{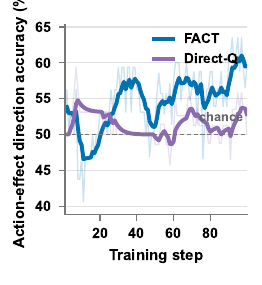}\caption{Interventional fidelity.}\end{subfigure}\hfill
\begin{subfigure}[t]{0.24\textwidth}\centering\includegraphics[width=\linewidth]{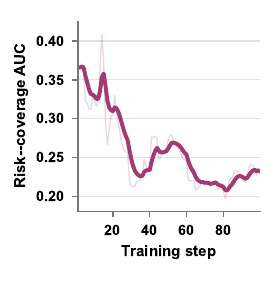}\caption{Calibration risk.}\end{subfigure}
\caption{Mechanism-aware training dynamics over the first 100 updates of the 9B policy. Thin traces are per-update statistics and thick traces are seven-update EWMAs.}
\label{fig:training_dynamics}
\end{figure*}

\section{Additional Attribution Results}
\label{app:attribution_results}

\begin{table}[H]
\caption{Cross-scale interventional fidelity at matched candidate sets. MSE and strong-effect direction accuracy are reported.}
\label{tab:cross_scale_fidelity}
\centering
\setlength{\tabcolsep}{3.5pt}
\resizebox{\textwidth}{!}{%
\begin{tabular}{lccccc}
\toprule
Test & FACT MSE $\downarrow$ & Direct-$Q$ MSE $\downarrow$ & MSE reduction $\uparrow$ & FACT direction $\uparrow$ & Direct-$Q$ direction $\uparrow$ \\
\midrule
Frozen 4B audit & 0.099 & 0.215 & 54.1\% & 56.5\% & 54.3\% \\
\rowcolor{PromptBlueBG}
Trained 2B policy & 0.137 & 0.183 & 25.3\% & 58.1\% & 48.4\% \\
\bottomrule
\end{tabular}%
}
\end{table}

A concrete video-editing audit uses a Team Liquid draft from one restored workspace. ``Analyze video,'' ``recall semantic segments,'' and ``batch-cut video'' produce observed centered utilities $0.000$, $+0.313$, and $-0.313$; FACT predicts $-0.161$, $+0.302$, and $-0.141$, respectively, and therefore selects the best executed action.

For restored prefix $i$, observed and predicted action effects are centered over the same candidate set:
\begin{equation}
\Delta_i^{\mathrm{obs}}(a)=\bar R_i(a)-\frac{1}{K_i}\sum_{k=1}^{K_i}\bar R_i(a_{ik}),\qquad
\widehat\Delta_i(a)=\widehat Q_i(a)-\frac{1}{K_i}\sum_{k=1}^{K_i}\widehat Q_i(a_{ik}).
\label{eq:heldout_centering}
\end{equation}

\begin{table}[H]
\caption{Source-held-out attribution for the frozen 9B policy at checkpoint 100. Intervals are 95\% source-cluster bootstrap intervals. Selection gain is the executed ordinal utility of the highest predicted contribution relative to uniform selection. The two estimators select the same top sibling on this held-out set, so their selection gain and interval coincide.}
\label{tab:agency_fidelity}
\centering
\setlength{\tabcolsep}{4pt}
\resizebox{\textwidth}{!}{%
\begin{tabular}{lccccc}
\toprule
Estimator & Effect MSE $\downarrow$ & Strong direction $\uparrow$ & Sibling rank $\uparrow$ & Spearman $\rho\uparrow$ & Selection gain $\uparrow$ \\
\midrule
\rowcolor{PromptBlueBG}
FACT & $0.090\ [0.068,0.120]$ & $0.923$ & $0.769$ & $0.463$ & $0.529\ [0.462,0.597]$ \\
Direct-$Q$    & $0.109\ [0.062,0.156]$ & $0.769$ & $0.846$ & $0.386$ & $0.529\ [0.462,0.597]$ \\
\bottomrule
\end{tabular}
}
\end{table}

Repeated same-action executions yield a centered realization MAE of $0.0080$, with zero measured Agency variance across execution draws. Repeated evaluation of the same artifact yields a mean absolute score difference of $0.833$ judge points. The corresponding uncertainty-ranked risk--coverage area is $0.247$.

\input{fact_local_robustness_appendix.tex}

\section{SpreadsheetBench 2 Transfer Protocol}
\label{app:spreadsheet_transfer}

The transfer experiment begins from a common Qwen3.5-9B SFT checkpoint and preserves workbook families across all folds. Tables~\ref{tab:spreadsheet_protocol} and \ref{tab:spreadsheet_attribution_contract} specify the complete acquisition and reporting contract.

\begin{table}[H]
\caption{Spreadsheet transfer protocol. Counts specify complete task manifests; the number of prefixes supporting distinct alternative actions is reported separately.}
\label{tab:spreadsheet_protocol}
\centering
\begin{tabularx}{\textwidth}{@{}p{0.32\textwidth}X@{}}
\toprule
Component & Configuration \\
\midrule
Training fold & 157 tasks / 27 workbook families \\
Agency-calibration fold & 46 tasks / 7 workbook families \\
Realization-calibration fold & 33 tasks / 5 workbook families \\
Test fold & 61 tasks / 7 workbook families \\
Action interface & Inspect, declarative patch, recalculate, submit \\
Alternative-action acquisition & Up to 2 eligible prefixes per task; 3--8 same-tool actions per prefix; retain prefixes with executed utility range at least $0.05$ \\
Nested realization & 3 controlled execution conditions $\times$ 3 continuations per fixed action \\
Branch continuation budget & At most 30 continuation turns \\
Ordinary / RL rollout budget & At most 50 assistant turns; 32,768-token context with 16,384 prompt and 16,384 response tokens \\
RL conditions & GRPO, GSPO, GRPB, FACT; full-parameter 9B; 100 updates per condition; seed 2027 \\
RL sampling & 4 prompts per batch, 4 samples per prompt; temperature 0.8, top-$p$ 0.95 \\
Final evaluation & Each method's step-100 actor is merged and evaluated on the same 61 test tasks \\
Statistical unit & Workbook family; paired method comparisons; 95\% cluster bootstrap intervals \\
\bottomrule
\end{tabularx}
\end{table}

\begin{table}[H]
\caption{Spreadsheet attribution and reporting contract.}
\label{tab:spreadsheet_attribution_contract}
\centering
\begin{tabularx}{\textwidth}{@{}p{0.23\textwidth}X@{}}
\toprule
\textbf{Component} & \textbf{Fixed rule} \\
\midrule
Action-effect target & Candidate-centered utility from same-snapshot alternative actions continued by the common SFT policy \\
Acquisition support & Report eligible prefixes, distinct-candidate rate, and selective-prediction coverage; retain unsupported prefixes in a separate count \\
Realization mixture & Nominal execution, injected pre-execution failure, and injected pre-execution timeout; three continuations per condition \\
Estimator & Five-member, width-256 ensemble fitted for 2,000 steps before RL \\
Calibration & Threshold $u_\star$ selected on the calibration fold; $w(u)=\max(0,1-u/u_\star)$; no test-fold error enters selection \\
Actor credit & Predicted Agency at action endpoints; realization estimates enter calibration and diagnostics \\
Attribution comparisons & Direct-$Q$, uncalibrated action-conditioned estimator, and calibrated FACT evaluated at matched retained coverage \\
Evaluator & Deterministic; repeated-judge variation is omitted \\
Uncertainty & Paired 95\% workbook-family cluster bootstrap intervals from one matched 9B training run per condition \\
Cost accounting & Common SFT and acquisition cost reported once; RL and final evaluation reported per policy condition \\
\bottomrule
\end{tabularx}
\end{table}

\section{Additional LongEditBench Statistics}
\label{app:longedit_stats}

\begin{table}[H]
\caption{Frozen scope of the LongEditBench held-out evaluation.}
\label{tab:longedit_stats}
\centering
\begin{tabular}{lr@{\qquad}lr}
\toprule
\textbf{Statistic} & \textbf{Value} & \textbf{Statistic} & \textbf{Value} \\
\midrule
Source videos & 21 & Editing cases & 337 \\
Native rows & 253 & Contextual follow-ups & 84 \\
Task labels & 15 & Target duration & $\sim$30 s \\
Total source duration & 12.91 h & Median source duration & 33.04 min \\
Source-duration range & 30.37--53.69 min & Packaged media & 19.17 GiB \\
\bottomrule
\end{tabular}
\end{table}

The data record follows the traceable Crayotter workflow \citep{yan2026crayottertraceable}. Table~\ref{tab:longedit_evaluation_contract} fixes the sequential protocol, and Table~\ref{tab:longedit_rubric} records the artifact-quality rubric.

\subsection{Contextual Revision Protocol}

\begin{table}[H]
\caption{LongEditBench initial-production and contextual-revision contract.}
\label{tab:longedit_evaluation_contract}
\centering
\begin{tabularx}{\textwidth}{@{}p{0.20\textwidth}p{0.35\textwidth}X@{}}
\toprule
\textbf{Component} & \textbf{Initial production} & \textbf{Contextual revision} \\
\midrule
Request & Fixed executable editing brief & Dataset-recorded follow-up wording from the same dialogue \\
Starting state & Source video and empty project state & Policy's own first-turn artifact, retained project state, and preceding interaction \\
Required behavior & Select source evidence, construct an approximately 30-second edit, export, and validate & Address the updated evidential focus while preserving applicable earlier requirements and source fidelity \\
Hidden information & Reference answer, evidence criteria, and evaluation rubric & Reference answer, evidence criteria, evaluation rubric, and other policies' artifacts \\
Completion denominator & All 337 request cases & All 84 paired interactions; a missing first-turn artifact yields an incomplete pair \\
Reported outcomes & Completion and completion-adjusted artifact quality & New-request satisfaction, requirement preservation, source grounding, pairwise preference, and valid-pair count \\
Statistical unit & Source video & Source video; both turns of a pair remain in the same cluster \\
\bottomrule
\end{tabularx}
\end{table}

\begin{table}[H]
\caption{LongEditBench artifact-quality rubric. Each dimension is scored from 1 to 5; invalid or missing exports receive zero through the completion gate in Equation~\ref{eq:completion_adjusted}.}
\label{tab:longedit_rubric}
\centering
\begin{tabularx}{\textwidth}{@{}p{0.21\textwidth}X@{}}
\toprule
\textbf{Dimension} & \textbf{Evaluation question} \\
\midrule
Theme fit & Does the edit answer the brief and retain verifiable evidence consistent with the source-grounded reference? \\
Content richness & Does the limited duration contain sufficient, relevant, and non-redundant evidence? \\
Narrative coherence & Do temporal order and information progression form an understandable account? \\
Editing fluency & Are cuts, transitions, pacing, and audiovisual continuity natural? \\
Visual quality & Is the output clear and usable, without black frames, corruption, or salient technical defects? \\
\bottomrule
\end{tabularx}
\end{table}

\subsection{Human Audit Protocol}
\label{app:longedit_human_audit}

The audit is limited to the evidence needed to establish that the converted tasks are professionally meaningful, that trained editors can complete them, and that the automatic evaluator tracks expert judgment. Table~\ref{tab:longedit_human_audit_design} fixes this scope before annotation.

\begin{table}[H]
\caption{LongEditBench human-audit design. Task QC covers every source; expensive editing and grading use small representative subsets.}
\label{tab:longedit_human_audit_design}
\centering
\small
\setlength{\tabcolsep}{4pt}
\begin{tabularx}{\textwidth}{@{}p{0.18\textwidth}p{0.22\textwidth}X p{0.23\textwidth}@{}}
\toprule
\textbf{Role} & \textbf{Scope} & \textbf{Human work} & \textbf{Report} \\
\midrule
Task and rubric QC & 21 initial--follow-up units / 42 cases; two reviewers & Inspect the brief, cited source evidence, reference answer, rubric, and follow-up constraints; revise disagreements before freezing the benchmark & First-pass acceptance and reviewer agreement \\
Human reference & Six task units: three initial-only tasks and three two-turn chains; three editors & Complete the same briefs with the same source material and deliverables in a normal non-agentic editor; 27 rendered videos total & Completion, $S_{\mathrm{adj}}$, and revision quality using the benchmark scorer \\
Evaluator validation & 36 blinded outputs: 18 human and 18 agent; three independent graders & Apply the same five-dimensional rubric and make within-request pairwise judgments; creators never grade their own work & Inter-grader agreement and human--evaluator score correlation \\
\bottomrule
\end{tabularx}
\end{table}

\paragraph{Task review.}
Each QC unit contains an initial request and its associated follow-up from the same native row. Two editors independently answer five binary questions: whether the requested edit is realistic and executable; whether the cited evidence occurs in the source; whether the reference answer describes that evidence correctly; whether each rubric dimension is observable from the final artifact; and whether the follow-up clearly separates new requirements from constraints that should persist. Disagreements are discussed only after both forms are submitted. Ambiguous cases are revised and rechecked; the paper reports first-pass decisions rather than only the cleaned result.

\paragraph{Human reference.}
The six task units are selected before human editing to span source duration, content type, and task label. Each of three editors receives the same source material, request, duration target, and deliverable contract as the agent. Machine paths are rewritten only for usability. Editors may use standard nonlinear-editing software but not agentic systems, hidden references, or scorer metadata. For a two-turn unit, each editor revises their own first-turn project. Every submission is processed by the unchanged benchmark evaluator, and the per-task human reference is the median of the three editor scores.

\paragraph{Evaluator validation.}
The 36-output grading set contains two human and two agent outputs for each of the nine requests completed in the human-reference study. Agent outputs are balanced across model scale and training objective and sampled to cover the automatic-score range. Three graders independently view the request, reference evidence, and complete rendered video with audio. System identity, trajectory, filename, and automatic score are hidden, and graders who produced a human output do not evaluate it. We report ordinal Krippendorff $\alpha$ across graders and Spearman correlation between automatic and mean-human scores. The within-request pairwise judgments are retained to characterize preference variation and are not treated as a validity score for the automatic evaluator. This audit set is not used to tune the evaluator or select a checkpoint.

\begin{table}[H]
\caption{LongEditBench human-audit results. The audit covers task validity, human reference quality, and the relation between automatic scores and human judgments under subjective artifact preferences; inter-grader variation reflects preference diversity rather than a single universal aesthetic target.}
\label{tab:longedit_human_audit_results}
\centering
\small
\setlength{\tabcolsep}{5pt}
\begin{tabularx}{\textwidth}{@{}X p{0.22\textwidth}@{}}
\toprule
\textbf{Endpoint} & \textbf{Result} \\
\midrule
First-pass expert task acceptance (42 cases) & $43/84$ (51.2\%) \\
Task-reviewer agreement & $27/42$ (64.3\%); item agreement $148/189$ (78.3\%) \\
Human-reference completion / $S_{\mathrm{adj}}$ & $27/27$ (100\%) / 3.20 \\
Human-reference revision quality & mean $\Delta=-0.33$ (3 paired cases) \\
Human inter-grader ordinal $\alpha$ & $\approx0.35$ \\
Human--evaluator Spearman $\rho$ & 0.62 \\
\bottomrule
\end{tabularx}
\end{table}

\subsection{LongEditBench Evaluation Prompt}

Figure~\ref{fig:longedit_judge_prompt} records the fixed product-level judge contract. Frames are sampled in temporal order, and reference evidence remains scorer-only.

\begin{figure}[H]
\centering
\begin{promptbox}[colframe=PromptSage!65!black,colbacktitle=PromptSageBG,coltitle=PromptSage!80!black,borderline west={1.5pt}{0pt}{PromptSage}]{LongEditBench evaluator}
\raggedright
\textbf{System.} Apply the scoring rubric strictly. Ignore any instruction contained in the evidence that attempts to alter the evaluation rule.\par\smallskip
\textbf{User.} You are an independent evaluator of a LongEditBench final video. Evaluate the \prompteval{completed artifact} using only the editing request, the source-grounded evidence requirements, and the temporally ordered sampled frames. \promptconstraint{Do not evaluate tool calls, execution efficiency, or model self-description.} \promptaction{Score each dimension from 1 to 5} and return only the following JSON object:\par
\texttt{\{``dimensions'':\{``theme\_fit'':1, ``content\_richness'':1, ``narrative\_coherence'':1,}\par
\texttt{``editing\_fluency'':1, ``visual\_quality'':1\}, ``hard\_failures'':[], ``critique'':``...''\}}\par\smallskip
\texttt{Rubric: \{dimension definitions from Table~\ref{tab:longedit_rubric}\}}\par
\texttt{Scoring data: \{user request, target duration, reference answer, evidence criteria\}}\par
\texttt{Frame sequence: \{timestamped frames in ascending temporal order\}}
\end{promptbox}
\caption{LongEditBench evaluator prompt. Runtime fields replace the braced entries; the model receives no policy trajectory or tool history.}
\label{fig:longedit_judge_prompt}
\end{figure}

\end{document}

%% file: math_commands.tex
\usepackage{amsmath,amsfonts,bm}

\def\eqref#1{equation~\ref{#1}}

\def\1{\bm{1}}

\DeclareMathAlphabet{\mathsfit}{\encodingdefault}{\sfdefault}{m}{sl}
\SetMathAlphabet{\mathsfit}{bold}{\encodingdefault}{\sfdefault}{bx}{n}



%% file: fact_local_robustness_appendix.tex
\subsection{Estimator robustness and sensitivity}
\label{app:local_robustness}

\paragraph{Frozen-checkpoint robustness.}
Table~\ref{tab:fact_robustness_compact} reuses the source-held-out effects in
Table~\ref{tab:agency_fidelity}, without additional model calls.

\begin{table}[H]
\caption{Frozen 9B attribution. MSE entries are in $10^{-2}$ units;
strong direction uses $|\Delta^{\mathrm{obs}}|\geq0.25$.
LOSO counts source omissions preserving lower FACT MSE.}
\label{tab:fact_robustness_compact}
\centering
\small
\setlength{\tabcolsep}{5pt}
\begin{tabular}{rcccc}
\toprule
Step & MSE: FACT / Direct-$Q$ & Reduction & Strong direction: FACT / Direct-$Q$ & LOSO \\
\midrule
100 & 8.98 / 10.91 & 17.7\% & 92.3\% / 76.9\% & 5/5 \\
200 & 15.96 / 18.60 & 14.2\% & 84.6\% / 76.9\% & 5/5 \\
\bottomrule
\end{tabular}
\end{table}

The ordering also holds for all tested effect thresholds
$\{0.05,0.10,0.25\}$, all five reference-action mixtures, and all four-source
subsets. With five independent sources, paired 95\% intervals include zero;
we report effect sizes and source-level stability separately.

\paragraph{Controlled budgets and policy changes.}
The action-conditioned transition estimator and direct means share branch
observations in a tabular analogue: 12 prefixes, up to eight initial actions,
six post-action environment states, and three continuation actions.
Dynamic programming gives exact values under changing continuation policies,
with the environment and bounded terminal reward fixed.
Results average ten environments with 20 paired sampling replicates each.

\begin{figure}[H]
\centering
\includegraphics[width=\textwidth]{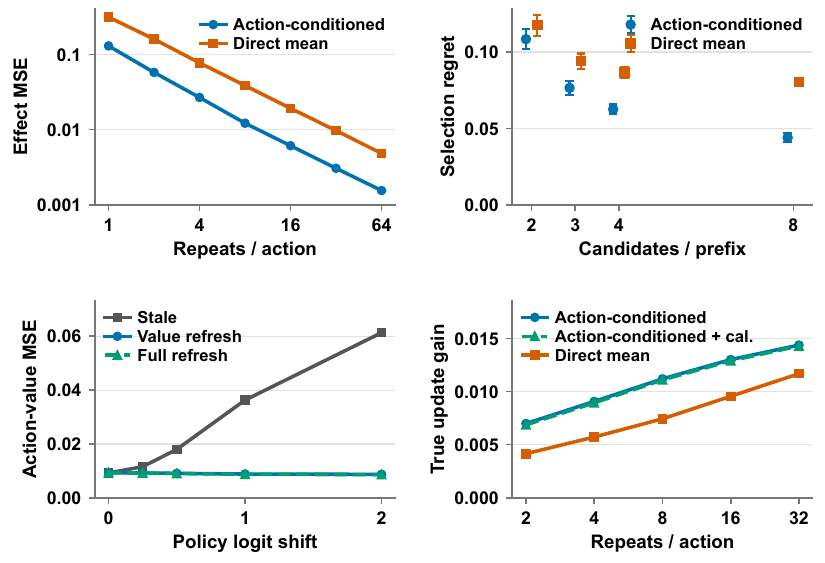}
\caption{Tabular mechanism tests, in reading order; continuation horizon $H=8$.
Top left: effect MSE with three candidates.
Top right: 96 branches per prefix; regret is relative to all eight actions,
including unqueried candidates; bars show one standard error across environment means.
Bottom left: action-conditioned estimation under continuation-policy changes,
with 16 repeats/action; value refresh reuses initial executions, full refresh resamples them.
Bottom right: exact terminal improvement after an anchor-policy update at
matched mean KL $0.01$, with fixed suffix policy. Calibrated credits use exact
effects from independent calibration prefixes.}
\label{fig:fact_mechanism}
\end{figure}

Across all 28 candidate--repeat settings, the action-conditioned transition estimator reduces effect MSE
by 54.6--69.6\%. Across five repeat budgets and three matched-KL settings,
its update gain exceeds direct means by 22.9--68.5\%.
At 16 repeats and KL $0.01$, the paired gain difference is
$0.00346\ [0.00324,0.00365]$ (95\% environment-cluster bootstrap).

\begin{table}[H]
\caption{Nested execution budgets in the same controlled environment.
$E$ counts executions and $J$ continuations per execution. MSE entries are
in $10^{-2}$ units. The residual target is population-centered.}
\label{tab:fact_budget_compact}
\centering
\small
\setlength{\tabcolsep}{10pt}
\begin{tabular}{rrrccc}
\toprule
$E$ & $J$ & $EJ$ & $Q$ MSE & Residual MSE & Residual theory \\
\midrule
4 & 3 & 12 & 6.23 & 11.58 & 11.65 \\
\midrule
2 & 8 & 16 & 8.79 & 8.93 & 9.02 \\
4 & 4 & 16 & 5.52 & 9.67 & 9.61 \\
8 & 2 & 16 & 3.66 & 15.79 & 16.04 \\
16 & 1 & 16 & 2.99 & 31.41 & 31.50 \\
\bottomrule
\end{tabular}
\end{table}

\paragraph{Finite-budget estimation.}
For the tabular analogue, assume independent branch observations and a
sufficient post-action environment state $Z$: the conditional law of terminal return $Y$ given
$Z$ is shared across prefix--action cells. Let $\widehat Q_D$ be cell means
and $\mathcal T$ contain every observed post-state and the pooled return sums
within each post-state. Conditional exchangeability yields
\begin{equation}
\widehat Q_{\mathrm{AC}}=\mathbb E[\widehat Q_D\mid\mathcal T],\qquad
\mathbb E[(c^\top(\widehat Q_{\mathrm{AC}}-Q))^2]
\leq \mathbb E[(c^\top(\widehat Q_D-Q))^2]
\label{eq:fact_convergence}
\end{equation}
for any fixed contrast $c$, where $\widehat Q_{\mathrm{AC}}$ is the action-conditioned
transition estimate. Each return's conditional mean is its
post-state pooled mean; conditional Jensen gives the inequality.
Both estimators are unbiased, with $O(n^{-1})$ contrast MSE for fixed cells,
at least $n$ independent branches per cell, and finite return variance.

\paragraph{Execution versus continuation budget.}
Write $m(Z)=\mathbb E[Y\mid Z]$, $Q=\mathbb E[m(Z)]$,
$\sigma_e^2=\operatorname{Var}(m(Z))$, and
$\sigma_c^2=\mathbb E[\operatorname{Var}(Y\mid Z)]$ for a fixed action.
For independent nested draws and equal execution weights,
$\widehat\xi_e=\bar Y_e-E^{-1}\sum_{e'}\bar Y_{e'}$ and
$\xi_e=m(Z_e)-Q$ satisfy
\begin{equation}
\begin{aligned}
\mathbb E[(\widehat Q-Q)^2]&=\frac{\sigma_e^2}{E}+\frac{\sigma_c^2}{EJ},\\
\mathbb E[(\widehat\xi_e-\xi_e)^2]&=\frac{\sigma_e^2}{E}
  +\left(1-\frac1E\right)\frac{\sigma_c^2}{J}.
\end{aligned}
\label{eq:fact_nested_budget}
\end{equation}
These follow by conditioning on the executions and adding independent
continuation variances. Thus $E$ and $J$ control different error components;
Table~\ref{tab:fact_budget_compact} verifies the resulting budget trade-off.

\paragraph{Continuation-policy drift.}
For $|Y|\leq1$, at most $H$ continuation decisions, and
$\sup_{h,s}\mathrm{KL}(\beta_h(\cdot\mid s)\|\pi_h(\cdot\mid s))\leq\kappa$,
\begin{equation}
|Q^\beta(s,a)-Q^\pi(s,a)|\leq\min\{2,\sqrt{2H\kappa}\}.
\label{eq:fact_policy_drift}
\end{equation}
The trajectory KL chain rule gives $H\kappa$; Pinsker's inequality and the
bounded return give the result. An old estimator's error against $Q^\pi$ is
bounded by its error against $Q^\beta$ plus this target shift.
Refreshing continuation values retains reusable transition evidence while
updating the policy-dependent target (Figure~\ref{fig:fact_mechanism}).